%% file: acl_latex.tex
\pdfoutput=1

\documentclass[11pt]{article}

\usepackage[]{acl}

\usepackage{times}
\usepackage{latexsym}
\usepackage{booktabs}
\usepackage{amsmath}
\usepackage{multirow}
	
\usepackage{soul}

\usepackage[T1]{fontenc}

\usepackage[utf8]{inputenc}

\usepackage{microtype}
\usepackage{graphicx}

\graphicspath{ {./images/} }

\title{Neural Retriever and Go Beyond: A Thesis Proposal}

\author{Man Luo \\
  Arizona State University \\
  \texttt{mluo26@asu.edu} \\
 }

\begin{document}
\maketitle

\begin{abstract}
Information Retriever (IR) aims to find the relevant documents (e.g. snippets, passages, and articles) to a given query at large scale. 
IR plays an important role in many tasks such as open domain question answering and dialogue systems, where external knowledge is needed.  In the past, searching algorithms based on term matching have been widely used. Recently, neural-based algorithms (termed as neural retrievers) have gained more attention which can mitigate the limitations of traditional methods. Regardless of the success achieved by neural retrievers, they still face many challenges, e.g. suffering  from a small amount of training data and failing to answer simple entity-centric questions. Furthermore, most of the existing neural retrievers are developed for pure-text query. This prevents them from handling multi-modality queries (i.e. the query is composed of textual description and images). 
This proposal has two goals. First, we introduce methods to address the abovementioned issues of neural retrievers from three angles, new model architectures, IR-oriented pretraining tasks, and generating large scale training data. Second, we identify the future research direction and propose potential corresponding solution\footnote{Since previous work use context, documents or knowledge to represent the retrieved information given a query, we use these two terms interchangeably.}. 
\end{abstract}

\section{Introduction}
The convenience and advance of internet not only speed up the spread of information and knowledge, but also the generation of new information. 
Such phenomenon also boosts humans needs of knowledge and frequency of acquiring information, which makes  
Information retrieval (IR) an important task in human life. 
IR aims to find relevant information from a large corpus to satisfy an information need.
It also plays an important role in other tasks such as open domain question answering and open domain dialogue, where external knowledge are needed. 
Not only that, IR can also assistant other systems to achieve a tough goal. By providing external knowledge, IR can help numerical reasoning systems to reach the correct answer~\cite{mishra2022numglue} , and IR can enrich or update the knowledge of large pretrained language models (PrLMs)~\cite{petroni2019language,sung2021language}.
By filtering and selecting examples~\cite{liu2021makes,lin2022unsupervised}, IR can assist in-context learning (ICL), a process allows large PrLMs do a new task instructed by prompts and few examples with few-shot tuning~\cite{gao2021making} or without any fine-tuning~\cite{brown2020gpt}.
\begin{figure*}[t]
    \centering
    \includegraphics[width=\linewidth]{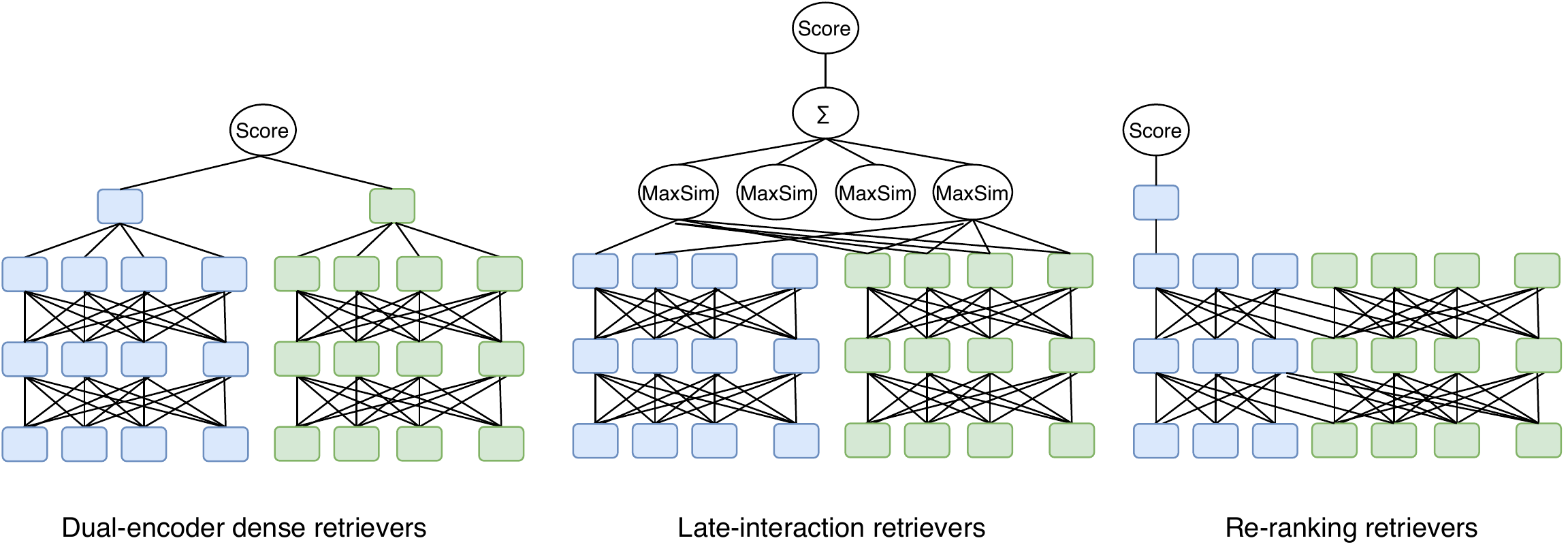}
    \caption{Architectures of three major types of retrievers. For simplicity, some lines in the figures are not drawn. Blue blocks represent the encoding for question, and the green blocks represent context or documents.}
    \label{fig:multi-retriever}
\end{figure*}

IR has a long history and the first automated information retrieval system can be traced back to the 1950s.
In this work, we call information retrieval  methods or systems as retrievers. 
Traditional retrievers are mainly based on term-matching, i.e. searching for information that has an overlap with terms in the query. 
TF-IDF and BM25~\citep{Robertson2009ThePR} are two strong and efficient algorithms in this category. 
Although these algorithms consider the importance and frequency of terms in query and document, they suffer from term-mismatch issues and lack of semantic understanding of the query and document~\cite{Chang2020PretrainingTF}.
Using neural models to represent the concatenation of query and passage is a promising way to achieve semantic matching~\citep{Nogueira2019PassageRW,Banerjee2020KnowledgeFA}.
These methods are only applicable at small scale retrieval but not at large scale. 
Recently, dual-encoder architecture retrievers based on large pretrained language models (PrLMs), such as BERT~\citep{devlin2019bert} have shown capability to do semantic matching and can be applicable at large scale~\citep{Karpukhin2020DensePR,guu2020realm,Lewis2020rag}.
Such neural retrievers (NR) involve two PrLMs which are used to compute the vector representation of queries and documents respectively. 
Neural retrievers are trained in such a way that the documents which best answer a query maximize the dot product between the two representations. 
Despite the success of neural retrievers, they still face many challenges. 
In the next Section, we will present a brief overview of five types of retrievers and the efforts made toward building stronger retrievers.
Section \ref{sec:limi_solu} describes four limitations of current NRs and promising solutions. 
Section \ref{sec:future_work} discusses three more research directions and potential solutions.
We conclude the proposal in Section \ref{sec:conc}.

\input{related_work}
\input{limitation_solution}

\input{future_work}

\section{Conclusion}\label{sec:conc}

In this proposal, we focus on an important task: information retrieval.
From word-matching retrievers to neural retrievers, many efforts have been made toward building  stronger retrievers that can achieve high recall and precision. 
We summarize five types of modern retrievers and methods to address some existing issues. 
While the development in this field is exciting, retrievers still have a long journey to go. 
We hope this proposal can shed some light on building a more capable retriever in future.  
\clearpage

\bibstyle{acl_natbib}
\bibliography{custom}

\end{document}

%% file: related_work.tex
\section{Retrievers in General} \label{sec:evolution_ir}
In general, the modern retrievers can be categorized in five classes (adapted from~\citep{Thakur2021BEIRAH}). 
\textbf{Lexical retrievers} such as BM25 are based on token-matching between two high-dimensional sparse vectors. The sparse vectors are represented based on the frequency of the terms in documents and thus does not require any annotated training data. Regardless of the simplicity of the algorithms, such methods perform well on new domains~\citep{Thakur2021BEIRAH}.
\textbf{Dual-encoder dense retrievers} consists of two encoders where the query encoder and context encoder generate  a single dense vector
representation for query and context respectively. 
Then the score can be computed by inner-dot product or cosine-similarity between the two representations~\citep{Karpukhin2020DensePR,xiong2020approximate,hofstatter2021efficiently}. Language models such as BERT~\cite{devlin2019bert} are preferred choices for encoders. 
\textbf{Sparse retrievers} use sparse representations instead of dense representations for query and document~\citep{dai2020context,zhao2021sparta,nogueira2019doc2query}. 
\textbf{Late-interaction retrievers} different from dense retrievers who use sequence-level representations of query and document, they use token-level representations for the query and passage: a bag of multiple contextualized token embeddings~\cite{khattab2020colbert}.
The late-interactions are aggregated with sum of the max-pooling query term and a dot-product across all passage terms.
\textbf{Re-ranking retrievers} include two stages, coarse-search by efficient methods (e.g. BM25) and fine-search by cross-attentional re-ranking models. 
The re-ranking model takes input as the concatenation of the query and one candidate given by the first stage and produce a score based on the cross representation (e.g. the [CLS] token), and such process is repeated for every candidate, and finally re-rank candidates based on the generated scores. 

Without changing the architectures, different efforts have been made toward learning better representation of dense vectors and improving the efficiency in terms of training resources as well as short inference time.
One way to improve the representation of dense vectors is to construct proper negative instances to train a neural retriever. 
In-batch negative training is a frequently used strategy to train dense retrievers, and the larger the batch size is, the better performance a dense retriever can achieve~\citep{Karpukhin2020DensePR,qu2021rocketqa}. 
Using hard negative candidates is better than using random or simple in-batch negative samples, for example, \citet{Karpukhin2020DensePR} mine negative candidates by BM25 and \citep{xiong2020approximate} mine negative candidates from the entire corpus using an optimized dense retriever. 
\citet{hofstatter2021efficiently} selects the negative candidates from the same topic cluster, such a balanced topic aware sampling method allows the training with small batch size and still achieves high quality dense representation. 
ColBert~\cite{khattab2020colbert} is proposed to improve the efficiency of the ranking model. Since every token can be pre-indexed, it prevents inference time from getting representation of context. 
While Colbert is faster than single-model, it is slower compared to dual-models, thus, it is not suitable for retrieval at large scale. 
On the other hand, \citet{nogueira2019doc2query} shortens the inference time by using sparse representation for queries. \citet{zhang2021adversarial} integrates dense passage retriever and cross-attention ranker and use adversarial training to jointly both module.  

Above methods are usually used to retrieve a document (e.g. a paragraph in Wikipedia) which can potentially contain the answer to a query. 
Some other retrievers directly retrieve the answer phrase (or entities) so that they can be directly used to answer questions without a reader~\cite{seo2019real,lee2020contextualized,de2020autoregressive,de2021multilingual}. 
While such methods can reduce the latency, it also increases the memory to store potential phrases which will be much larger than the number of raw documents. 
On the other hand, \citet{lee2021learning,lee2021phrase} use generative model to generate the entities which largely reduce the memory. 

%% file: limitation_solution.tex
\section{Research Gaps and Solutions}\label{sec:limi_solu}

In this section, we will describe multiple research gaps and the proposed methods introduced in \citep{luo2021just,luo2021weakly,luo2022improving}. 
\begin{figure}[h]
    \centering
    \includegraphics[width=\linewidth]{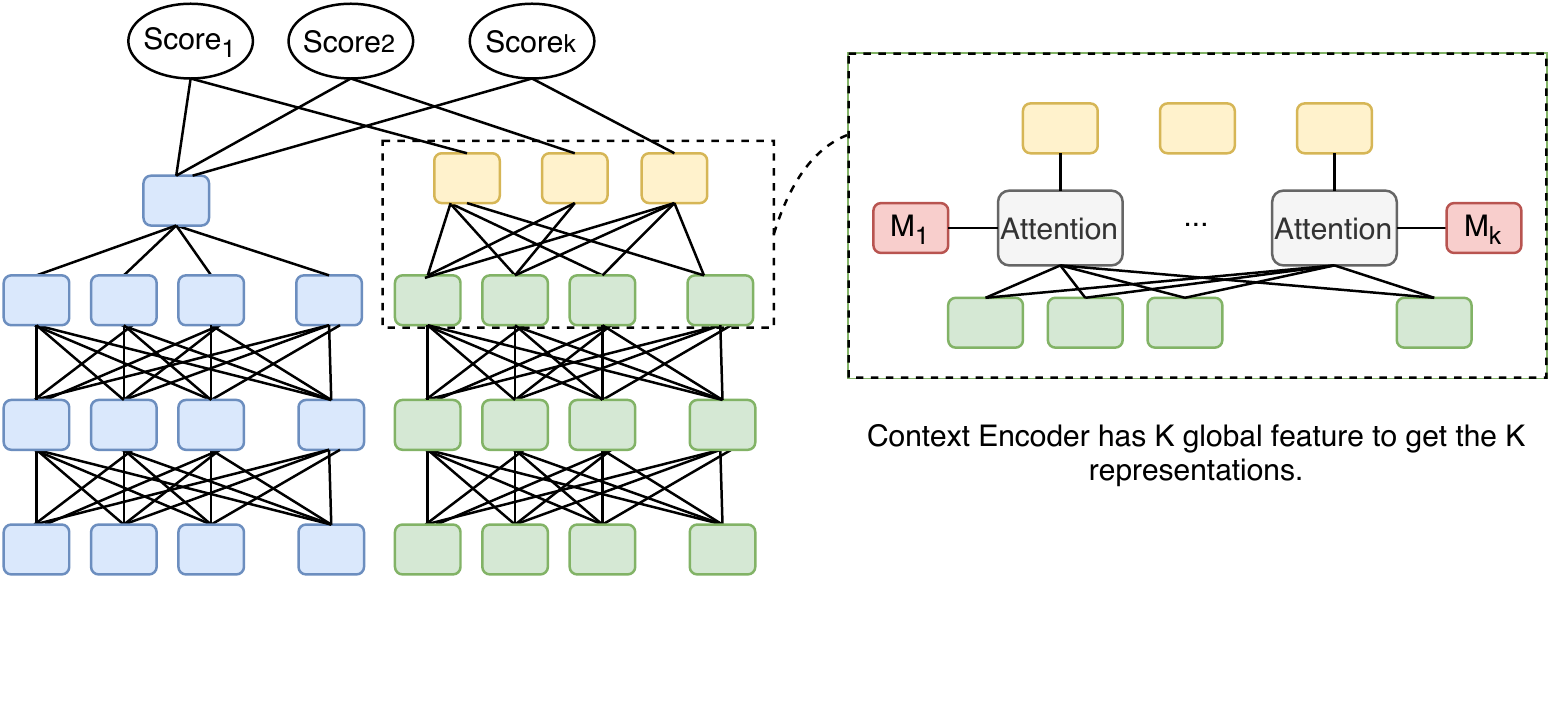}
    \caption{Poly-DPR, the context encoder uses K representations to capture the information in context.}
    \label{fig:poly-dpr}
\end{figure}
\begin{figure*}[t]
    \centering
    \includegraphics[width=0.90\linewidth]{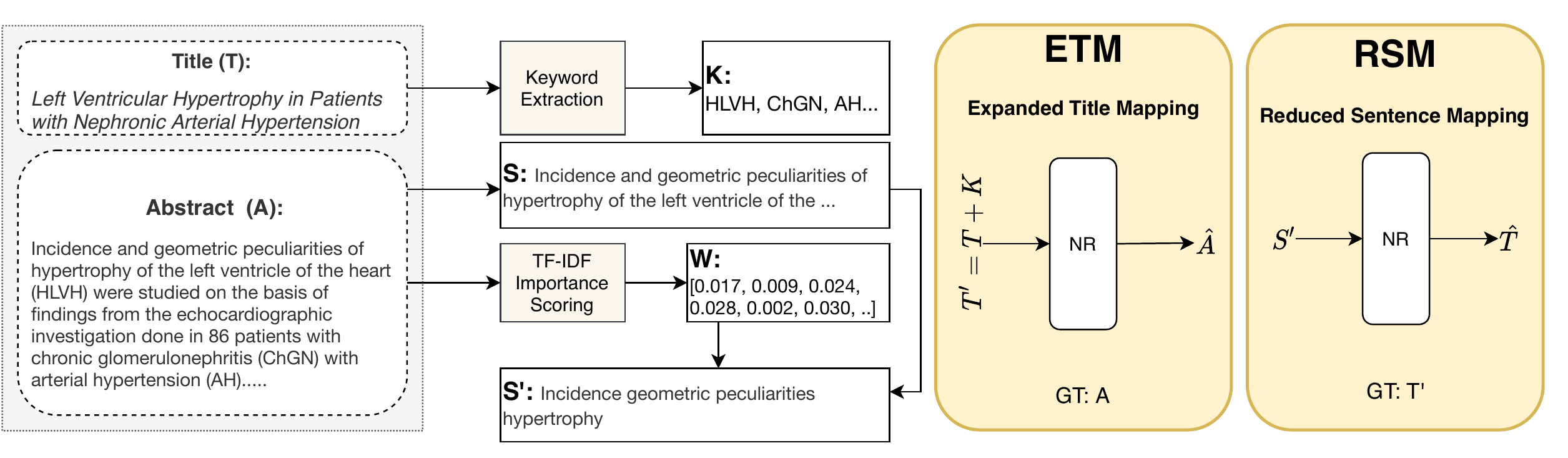}
    \caption{Two IR-oriented pretraining tasks. ETM is suitable for corpus which have titles and passages. RSM is suitable for any type of corpus.}
    \label{fig:pre-training}
\end{figure*}

\subsection{Is One Dense Vector Enough to Capture Information?}\label{sec:poly-dpr}

Most of the neural retrievers use one dense representation for context \citep{Karpukhin2020DensePR,guu2020realm,Lewis2020rag}.
Previous work found that one dense vector is not enough to capture enough information in the context, especially for a long context. One dense representation is also hard to be applied to exact word matching so that it fails on entities-centric questions~\citep{sciavolino2021simple}. 
To close the gap of existing NRs, we propose a new model called Poly-DPR which builds upon two recent developments: Poly-Encoder~\cite{Humeau2020PolyencodersAA} and Dense Passage Retriever~\citep{Karpukhin2020DensePR}. 

\paragraph{Method} 
In Poly-DPR (see Figure \ref{fig:poly-dpr}), the context encoder represents each \textit{context} using K vectors and produces \textit{query-specific vectors} for each context. 
In particular, the context encoder includes K global features $(m_1, m_2, \cdots, m_k)$, which are used to extract representation $v_c^i,~\forall i\in\{1 \cdots k\}$ by attending over all context tokens vectors. 
\begin{align}
    v_c^i &= \sum_{n} w_n^{m_i} h_n,~~\text{where}\\
    (w_1^{m_i} \dots, w_n^{m_i} ) &= \mathrm{softmax}(m_i^{T} \cdot h_1,  \dots, m_i^{T}\cdot h_n).
\end{align}

After extracting K representations, a query-specific context representation  $v_{c,q}$ is computed by using the attention mechanism: 
\begin{align}
    v_{c,q} &= \sum_k w_k v_c^k,~~\text{where}\\
    (w_1, \dots, w_k) &= \mathrm{softmax}(v_q^{T}\cdot v_c^1, \dots, v_q^{T}\cdot v_c^k).
\end{align}

To enable efficient search in inference (e.g. using MIPS~\cite{Shrivastava2014AsymmetricL} algorithms), instead of computing query-specific context representation, we simply use the inner-dot product of each K representations with the query embeddings, and apply max pooling function to get the score.

\paragraph{Result} We evaluate Poly-DPR on BioASQ8~\citep{nentidis2020overview} dataset to see how effective the model is. Instead of using the full corpus which has 19M PubMed articles, we construct a small corpus with 133,084 articles for efficient and comprehensive experiments purpose. 
We also examine the impact of changing the  value of K on the performance. Furthermore, we design two context length, one is two sentences no more than 128 tokens (short) and the other one is up to 256 tokens (long). In Table \ref{tab:kvalue}, we have three values for K, where value 0 is the same as the original DPR.  
We see that in both settings, Poly-DPR is better than the original DPR, and a larger value of K leads to better performance. 
\input{tables/exp_dpr_k}


\subsection{Is IR-oriented Pretraining Important?}

PrLMs are trained on general tasks, such as masked language prediction, and next sentence prediction~\citep{devlin2019bert}. 
While these pretraining tasks help the model to learn the linguistic knowledge, the model might still lack of specific skill to perform down-stream tasks, e.g. match similar words or characterize the relation between the question and answer.
\citet{Chang2020PretrainingTF} has shown that IR-oriented pretraining tasks can help model to develop basic retrieval skill. However, their proposed methods require specific document structure, e.g. the document includes external hyperlinks. 

\paragraph{Method} 
We propose two new IR-oriented pre-training strategies (Figure \ref{fig:pre-training}).
Our pre-training tasks are designed such that they can be used both for long contexts as short contexts. 
In \textbf{Expanded Title Mapping (ETM)}, the model is trained to retrieve an abstract, given an extended title $T^\prime$ as a query.
$T^\prime$ is obtained by extracting top-$m$ keywords from the abstract based on the TF-IDF score, denoted as $K = \{k_1, k_2, \cdots, k_m\}$, and concatenating them with the title as: $T' = \{T, k_1, k_2, \cdots, k_m\}$. 
The intuition behind ETM is to train the model to match the main topic of a document (keywords and title) with the entire abstract.
\textbf{Reduced Sentence Mapping (RSM)} is designed to train the model to map a sentence from an abstract with the extended title $T^\prime$.
For a sentence $S$ from the abstract, we first get the weight of each word $W = \{ w_1, w_2, \cdots, w_n\}$ by the normalization of TF-IDF scores of each word.  
We then reduce $S$ to $S'$ by selecting the words with the top-$m$ corresponding weights.
The intuition behind a reduced sentence is to simulate a real query which usually is shorter than a sentence in an abstract. 
\paragraph{Result} We test on BioASQ dataset and use the similar experimental setting as in \S\ref{sec:poly-dpr}, where we use both short and long context length settings. 
From Table \ref{tab:pt_ablation}, we see that in both settings, using our pretraining tasks are much better than without any pretraining with large margins. 
Furthermore, in the long context setting, we also compare our method with ICT~\cite{Lee2019LatentRF} pretraining task, and we see that ETM beats than ICT on average with better performance on 4 out of 5 batches. 
\input{tables/pt_tasks}

\subsection{How to Obtain Enough Training Data?}

While the pretraining makes language models more easily adapted to new tasks, a decent amount of domain-specific data for fine-tuning is still crucial to achieve good performance on downstream tasks~\citep{howard2018universal,clark2019electra}. 
Collecting annotated data is expensive and time consuming. 
Moreover, for some domains such as biomedical, annotation usually requires expert knowledge which makes the data collection harder~\cite{tsatsaronis2012bioasq}. 
To address this problem, \citet{ma2021zero} uses a question generation model trained on existing large scale data to obtain synthetic question-answer pairs using domain articles. Still, the style of the generated questions are far away from the target-domain and limit the models' performance. 

\begin{figure}[t]
    \centering
    \includegraphics[width=0.85\linewidth]{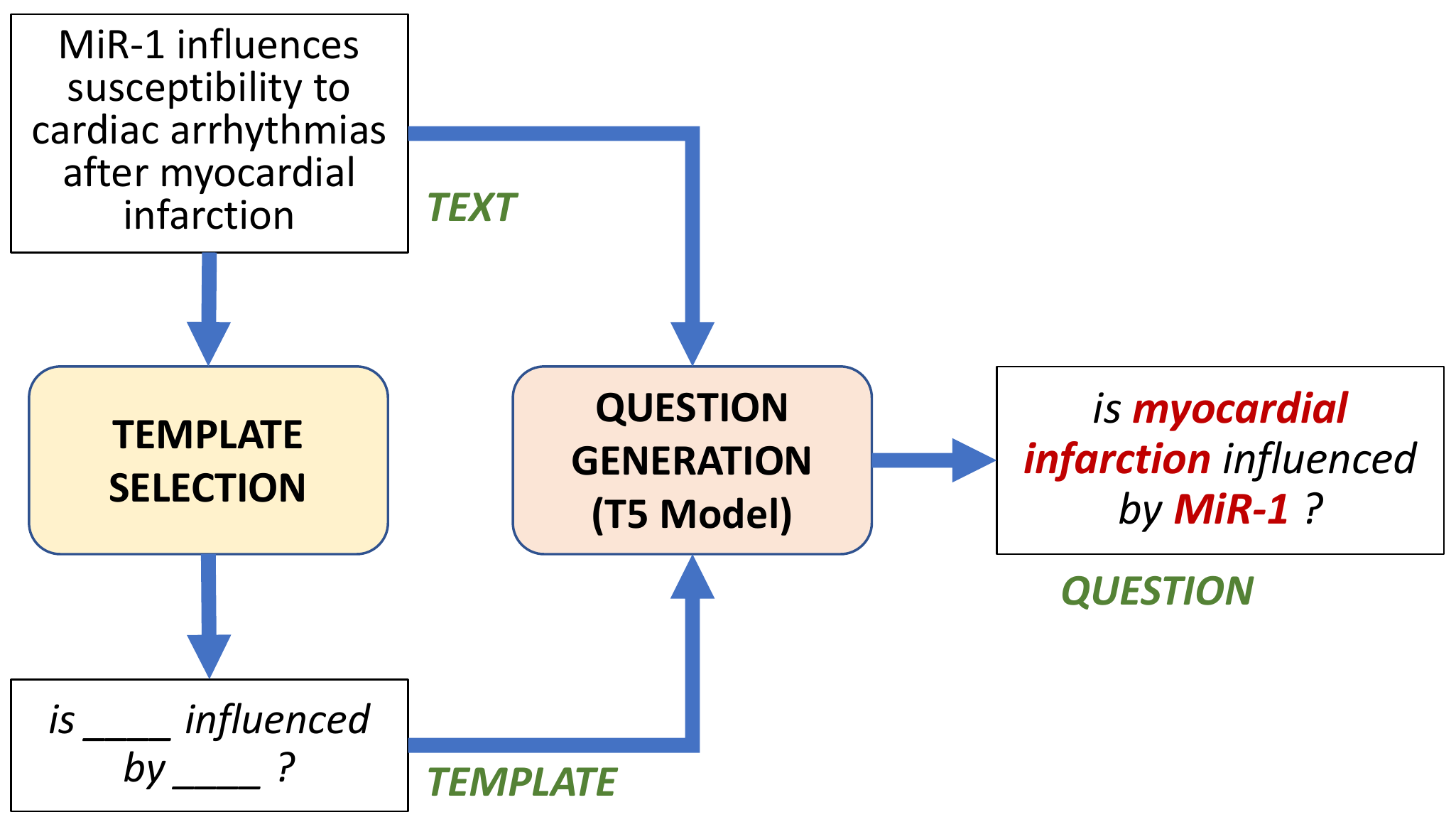}
    \caption{Template-Based Question Generation.}
    \label{fig:bionr_qg}
\end{figure}

\paragraph{Method}
To address the domain adaptation issue, we propose a semi-supervised pipeline to generate questions using domain-templates (Figure \ref{fig:bionr_qg}).
To do so, we assume a small amount of domain annotated question-answer data is given. We first extract templates from the questions by using a name entity recognition model to identify question-specific entities and removing such entities. 
A template selection model is trained to select the template for a new passage. Finally a generative model (e.g. T5) is trained to generate questions conditioned on this template and a text passage. 
The questions generated using domain templates are much better than the previous question generation method.

\paragraph{Result} 
Again, we use BioASQ8 as testbed with similar settings as previous experiments. 
We compare our method with an existing question generation method which extracts answer span first and then generates questions~\cite{Chan2019ARB}.
In Table \ref{tab:tempqg}, we compare three models trained on two generated questions as well as the training dataset of BioASQ8, and our proposed method is better than the other two especially with large gain (10\%+) in long context setting. 
\input{tables/tempqg}

\subsection{How to Retrieve Information for Multi-modality Queries?} \label{sec:multi_modality_retriever}
Previous discussion focuses on retrieving relevant documents to text-only queries, while in current society, lots of  information is presented by multi-modalities such as text, image, speech, and video. Therefore, retrieving relevant documents to multi-modality queries can have wide application in human's life. 
For instance an image of a milkshake and a complementary textual description ``restaurants near me'' should return potential matches of nearby restaurants serving milkshakes.  
In literature, OK-VQA~\citep{okvqa} is a task that requires external knowledge to answer visual questions (i.e. the query is composed of image and text.). To find the relevant knowledge for such a query, current neural retrieval can not be directly applied since the text part in the query is not completed to understand the information needs and the model is unable to look at the image information. To address this issue, we propose three types of retrievers to handle multi-modality queries.

\begin{figure}[t]
    \centering
    \includegraphics[width=0.9\linewidth]{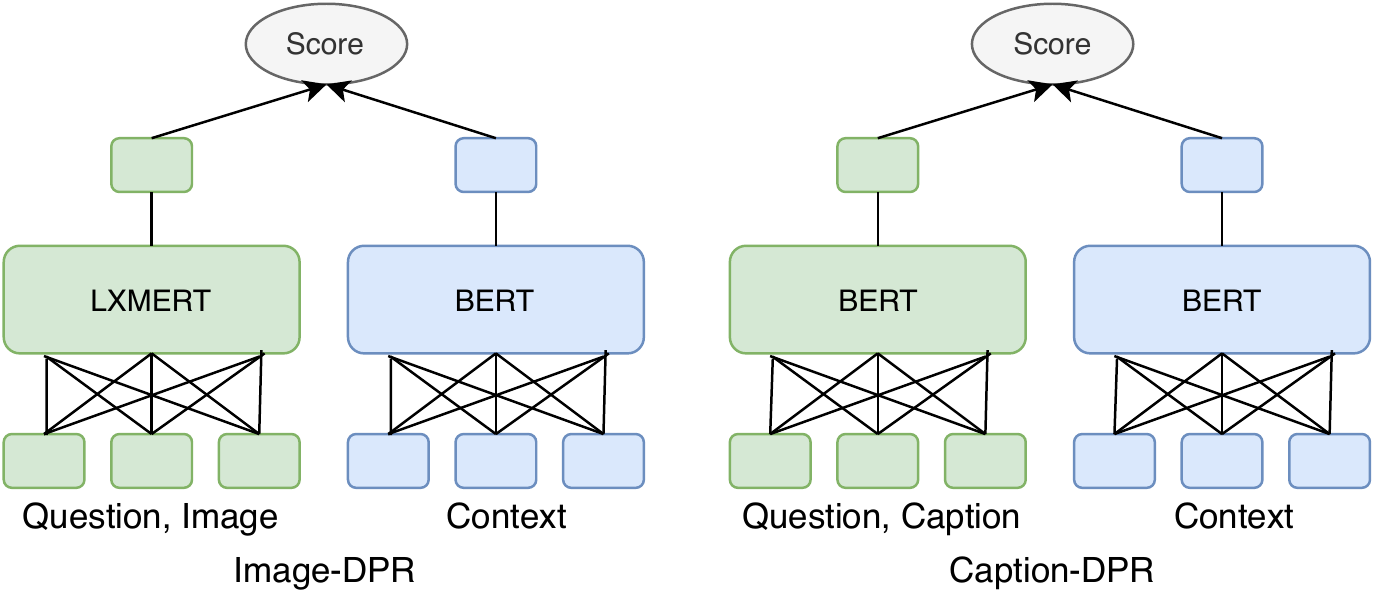}
    \caption{Comparison of two multi-modality. }
    \label{fig:mutli_modality_retriever}
\end{figure}

\paragraph{Method}\textit{Term-based retriever}, we first extract the image information by using a captions generation model~\citep{li2020oscar}. Then we concatenate
the question and the caption as a query and obtain knowledge by BM25. 
The other two multi-modality retrievers are adopted from the DPR model. 
\textit{Image-DPR}: we use LXMERT~\citep{Tan2019LXMERTLC} as the question encoder, which takes image and question as input and outputs a cross-modal representation.  
\textit{Caption-DPR}: similar to the strategy we use in term-based retrievers, we concatenate the question with the caption of an image as a query and use standard BERT as a query encoder to get the representation. 
In both \textit{Image-DPR} and \textit{Caption-DPR}, we use standard BERT as context encoder. 
Figure \ref{fig:mutli_modality_retriever} shows a comparison between these two retrievers. 
We find that the performance of Caption-DPR is better than Image-DPR, and the term-based retriever performs worst. 

\paragraph{Result}
\input{tables/okvqa}
We evaluate three retrievers on OK-VQA dataset and use the knowledge base (with 112,724 pieces of knowledge) created in \cite{luo2021weakly} as the corpus. 
We retrieve 1/5/10/20/50/80/100 knowledge for each question.
Table \ref{tab:retriever-eval} shows that the two neural retrievers are better than simple term-based retriever, and the Caption-DPR is the best model in all cases.



%% file: tables/exp_dpr_k.tex
\begin{table}[t]
    \small
    \centering
    \setlength{\tabcolsep}{3pt}
    \begin{tabular}{@{}cccccccc@{}}
        \toprule
         {\bf CL} & {\bf K} & {\bf B1}   & {\bf B2} & {\bf B3} & {\bf B4} & {\bf B5} & {\bf Avg.}\\
         \toprule
            \multirow{3}{*}{Short} & 0  & 62.06  & {\bf 61.81}& 61.85 & 66.69 & 61.30 & 62.74\\
            & 6  & 62.92 & 58.79 & {\bf 62.94} & 70.30 & 63.39 & 63.67\\
            & 12  & {\bf 65.22} & 60.86 & 62.59 & {\bf 70.50} & {\bf 66.21} & {\bf 65.08}\\
            \midrule
            \multirow{3}{*}{Long}  & 0  & 61.70  & 58.28 & 58.62 & 67.33 & 61.48 & 61.48\\
            & 6  & {\bf 63.95} & {\bf 59.51} & {\bf 62.98} & 66.71 & 62.80 & 63.19\\
            & 12 & 63.83 & 57.81 & 62.72 & {\bf 70.00} & {\bf 63.64} & {\bf 63.60}\\
        \toprule
    \end{tabular}
\caption{Comparison among different values of K for Poly-DPR in both short and long context settings of BioASQ8 dataset using MRR metric. B$i$ stand for different testing batch.}
\label{tab:kvalue}
\end{table}

%% file: tables/pt_tasks.tex
\begin{table}[t]
    \centering
    \small
    \setlength{\tabcolsep}{3pt}
    \begin{tabular}{@{}cccccccc@{}}
        \toprule
        \textbf{CL} & \textbf{PT} & \textbf{B1} & \textbf{B2} & \textbf{B3} & \textbf{B4} & \textbf{B5} & \textbf{Avg.}\\
        \toprule
    \multirow{2}{*}{\rotatebox[origin=c]{90}{Short}} 
    & -             &54.48      &50.51       & 53.8       &59.06     &48.71     & 53.31 \\
    & RSM           &\textbf{65.94}      &\textbf{57.43}       & \textbf{61.89}    &\textbf{69.01}    &\textbf{58.23 }     & \textbf{62.50}  \\
    \midrule
    \multirow{4}{*}{\rotatebox[origin=c]{90}{Long}}
    & -  &35.69 &32.66  &32.26 &38.28  &30.87 & 33.95  \\
    & ICT &54.44  &\textbf{47.37}  &52.61 &53.69  &44.38 &50.50  \\
    & ETM  & \textbf{56.63} &46.63 &\textbf{52.79} &\textbf{56.97}  &\textbf{49.61} &\textbf{52.53}  \\
    \bottomrule
    \end{tabular}
    \caption{Effect of pre-training tasks (PT) on the performance of Poly-DPR with two context lengths (CL) on the BioASQ dataset. 
    }
    \label{tab:pt_ablation} 
\end{table}
        

%% file: tables/tempqg.tex
\begin{table}[t]
    \centering
    \small
    \resizebox{\linewidth}{!}{
    \begin{tabular}{@{}ccccccccc@{}}
        \toprule
        \textbf{CL} & \textbf{PT} & \textbf{FT} & \textbf{B1} & \textbf{B2} & \textbf{B3} & \textbf{B4} & \textbf{B5} & \textbf{Avg.}\\
        \toprule
    \multirow{3}{*}{\rotatebox[origin=c]{90}{Short}} 
    & RSM      & B       &\textbf{65.94}      &57.43       & 61.89     &69.01    &58.23      & 62.50  \\
    & RSM & A & 56.84 & 55.79 & 57.52 & 58.68 & 55.15 & 56.80\\
    & RSM & T  & 64.71 & \textbf{64.92} & \textbf{64.28}  & \textbf{73.11}  &  \textbf{66.29} & \textbf{66.66} \\
    \midrule
    \multirow{3}{*}{\rotatebox[origin=c]{90}{Long}}
    & ETM & B & 56.63 &46.63 &52.79 &56.97  &49.61 &52.53  \\
    & ETM & A  & 54.44 & 49.95 & 48.42 & 58.15 & 52.60 & 52.71\\
    & ETM & T & 64.57 &58.51 & \textbf{64.02}  &68.44 &62.60 & \textbf{63.62}  \\
    \bottomrule
    \end{tabular}
    }
    \caption{Comparison of fine-tuning on different downstream training data B: BioASQ A: AnsQG and T: TempQG) on the performance of Poly-DPR with two context lengths (CL) on the BioASQ small corpus test set. 
    }
    \label{tab:tempqg} 
\end{table}

%% file: tables/okvqa.tex
\begin{table*}[ht!]
    \centering
    
    \resizebox{\textwidth}{!}{
    \begin{tabular}{lcccccccccccccc}
    \toprule
    \multirow{3}{*}{\bf Model}   &\multicolumn{12}{c}{\bf \# of Retrieved Knowledge }\\
    \cmidrule(lr){2-15}
    ~& \multicolumn{2}{c}{1} & \multicolumn{2}{c}{5}   &\multicolumn{2}{c}{10}  & \multicolumn{2}{c}{20}  & \multicolumn{2}{c}{50}  & \multicolumn{2}{c}{80} & \multicolumn{2}{c}{100} \\
    \cmidrule(lr){2-3}\cmidrule(lr){4-5}\cmidrule(lr){6-7}\cmidrule(lr){8-9}\cmidrule(lr){10-11}\cmidrule(lr){12-13}\cmidrule(lr){14-15}
    ~ &  P* & R* &  P* & R* &  P* & R* &  P* & R* &  P* & R* &  P* & R* &  P* & R* \\
    \toprule
    BM25 & 37.63 & 37.63 &  35.21 & 56.72 &  34.03 & 67.02 & 32.62 & 75.90 &  29.99 & 84.56 & 28.46 & 88.21 & 27.69 & 89.91 \\
    Image-DPR &  33.04 & 33.04 & 31.80 & 62.52 &  31.09 & 73.96 &  30.25 & 83.04 &  28.55 & 90.84 &  27.40 & 93.80 & 26.75 & 94.67 \\
    Caption-DPR & {\bf 41.62} & {\bf 41.62} &  {\bf 39.42} & {\bf 71.52} &  {\bf 37.94} & {\bf 81.51} &  {\bf 36.10} & {\bf 88.57} &  {\bf 32.94} & {\bf 94.13} & {\bf 31.05} &  {\bf 96.20}  &{\bf 30.01} & {\bf 96.95}\\
    \toprule
    \end{tabular}
    }
    \caption{
    Evaluation of three proposed visual retrievers on Precision (P) and Recall (R):
    Caption-DPR achieves the highest Precision and Recall on all number of retrieved knowledge. 
    }
    \label{tab:retriever-eval}
\end{table*}

%% file: future_work.tex
\section{Future Work}\label{sec:future_work}

Previous section describes multiple research problems for neural retrievers, while we provide some solutions, each problem can be further investigated. 
In the following, we identify more research directions and propose potential solutions. 

\paragraph{Document Expansion} 
Previous work~\citep{nogueira2019doc2query} has shown BM25 with expended documents using generated questions is an efficient way to retrieve documents. 
Such a method also showed good generalization across different domains~\citep{Thakur2021BEIRAH}. 
The template-based question generation proposed in this work has better domain adaptation than the previous question generation method. 
It is interesting to see how each module in the pipeline performs on new domain without further fine-tuning. For example, can the template selection model select good templates for passage from new domain; can the question generation model generate good questions given a new template? 
Evaluating how our template-based question generation pipeline works when apply it to document expansion is an interesting future work.

\paragraph{Distinguish Between Negative Samples}
Many training data only provide positive candidates but not the negative candidates. 
Section \ref{sec:evolution_ir} summarizes existing methods to construct negative candidates; however, the negativeness of different candidates are different.
For instance, if some candidates have the same topic as the queries while others do not, then in such cases, the former candidates should be less negative compared to the later. 
We propose to label the negativeness of candidates by using the similarity between the questions and the candidates and use such labels to train neural retrievers. 


\paragraph{Generalization of Neural IR}
Previous work has shown that neural retrievers perform well on the same domain of the training data (IID) but poorly in out-of-domain~\cite{Thakur2021BEIRAH}. 
In fact, generalization is a common issue in many other tasks such as image classification and question answering~\cite{gokhale2022generalized,luo2022choose}.
A range of methods including data augmentation, data filtering, and data debiasing methods have been proposed to improve the generalization capacity of models. 
Applying these methods to train neural retrievers can potentially improve their generalization capacity. 
Prompting or instruction learning has shown good generalization performance on many NLP tasks~\cite{mishra2021cross} or in low-resource domain~\cite{parmar2022boxbart}, yet applying such method  on retrieval task is less investigated, and it will be an interesting direction to explore.